\documentclass[letterpaper, 10 pt, conference]{ieeeconf}  

\IEEEoverridecommandlockouts                              

\usepackage{graphicx} 
\usepackage{amsmath} 
\usepackage{amssymb}  
\usepackage[caption=false]{subfig}
\usepackage{multicol}
\usepackage{booktabs}
\usepackage{color,soul}
\usepackage{fixltx2e}
\usepackage[ruled,vlined]{algorithm2e}
\usepackage{hyperref}
\usepackage{forloop}
\SetKwInput{KwInput}{Input}
\SetKwInput{KwOutputput}{Output}
\usepackage{makecell}

\title{
URA*: Uncertainty-aware Path Planning using Image-based Aerial-to-Ground Traversability Estimation for Off-road Environments
}

\author{Charles Moore$^{1}$, Shaswata Mitra$^{1}$, Nisha Pillai$^{1}$, Marc Moore$^{1}$, Sudip Mittal$^{1}$, Cindy Bethel$^{1}$, and Jingdao Chen$^{1}$
\thanks{$^{1}$ Computer Science and Engineering, Mississippi State University, Mississippi State, MS 39762, USA. Email: cam1271@msstate.edu, sm3843@msstate.edu, pillai@cse.msstate.edu, mnm419@msstate.edu,  mittal@cse.msstate.edu, cbethel@cse.msstate.edu, chenjingdao@cse.msstate.edu}
}

\begin{document}
\newcommand{\mysubscript}[1]{\raisebox{-0.34ex}{\scriptsize#1}}

\maketitle
\begin{abstract}
A major challenge with off-road autonomous navigation is the lack of maps or road markings that can be used to plan a path for autonomous robots. Classical path planning methods mostly assume a perfectly known environment without accounting for the inherent perception and sensing uncertainty from detecting terrain and obstacles in off-road environments. Recent work in computer vision and deep neural networks has advanced the capability of terrain traversability segmentation from raw images; however, the feasibility of using these noisy segmentation maps for navigation and path planning has not been adequately explored. To address this problem, this research proposes an uncertainty-aware path planning method, URA* using aerial images for autonomous navigation in off-road environments. An ensemble convolutional neural network (CNN) model is first used to perform pixel-level traversability estimation from aerial images of the region of interest. The traversability predictions are represented as a grid of traversal probability values. An uncertainty-aware planner is then applied to compute the best path from a start point to a goal point given these noisy traversal probability estimates. The proposed planner also incorporates replanning techniques to allow rapid replanning during online robot operation. The proposed method is evaluated on the Massachusetts Road Dataset, the DeepGlobe dataset, as well as a dataset of aerial images from off-road proving grounds at Mississippi State University. Results show that the proposed image segmentation and planning methods outperform conventional planning algorithms in terms of the quality and feasibility of the initial path, as well as the quality of replanned paths.
\end{abstract}

\section{INTRODUCTION}


A key step in navigating ground robots in unmapped, off-road environments is performing traversability estimation. The concept of traversability estimation refers to interpreting the geometry and appearance of the region of interest to determine whether a vehicle could drive through it safely depending on its capabilities~\cite{borges2022}~\cite{sharma2022}. In structured urban environments with clear road markings, local traversability estimation using sensors from a ground vehicle's perspective~\cite{levi2015}\cite{oliveira2016} is usually sufficient for navigation. Whereas in unstructured off-road environments such as dense forests or mountainous regions where the robot's field of view is limited, aerial-to-ground traversability estimation is advantageous to enable path planning from a global perspective~\cite{chavez2018}\cite{kim2019}\cite{hudjakov2009}.

Major advances in computer vision and deep neural networks (DNNs) have enabled work in traversability estimation from aerial images in the form of road segmentation~\cite{bandara2021}\cite{quan2021} or terrain segmentation~\cite{hosseinpoor2021}. However, these works only consider the traversability prediction task without addressing the path planning task, which needs to account for errors and uncertainty in the perception model output. Another line of work has proposed ad-hoc modifications to conventional path planners such as Rapidly-exploring Random Trees (RRT)~\cite{lavalle1998} and A*~\cite{hart1968} by adding terrain and slip penalization terms to make the planner more risk-aware~\cite{ono2015}\cite{candela2022}. However, in these studies, such penalization terms are usually hand-engineered from prior knowledge instead of using a traversability measure that can be directly obtained from sensor data. Thus, research gaps remain in consolidating robotic path planning algorithms with recent advances in learning-based traversability estimation techniques.

This research proposes an uncertainty-aware path planning algorithm using aerial traversability estimation for off-road environments. An ensemble convolutional neural network (CNN) model is first used to perform segmentation of aerial images and output a traversal probability value at the pixel level. Given the noisy traversal probability estimates, an uncertainty-aware path planning algorithm is proposed to predict the best global path for a ground robot to travel from its start location to the goal location. A probabilistic replanning technique that combines information from noisy aerial-to-ground traversability estimates with accurate ground-level traversability measurements is applied so that the ground robot is able to rapidly scan and re-plan suitable paths during physical operation.
Code \footnote[1]{\url{github.com/shaswata09/Offroad-Path-Planning/}} and datasets \footnote[2]{\url{kaggle.com/datasets/mitrashaswata/msstate-cavs-off-road-aerial-images}} are made publicly-available.  

In summary, the key contributions of this work are: 
\begin{itemize}
    \item Development of an uncertainty-aware global path planning algorithm that makes use of noisy aerial-to-ground traversability estimates, with a strong coupling between perception and planning.
    \item Introduction of a probabilistic replanning technique to enable online updates of the planned path.
    \item Demonstrated the feasibility of this path-planning approach through experiments with three different image datasets, including a challenging off-road environment.
\end{itemize}

\begin{figure*}[]
  \centering
  \includegraphics[scale=0.6]{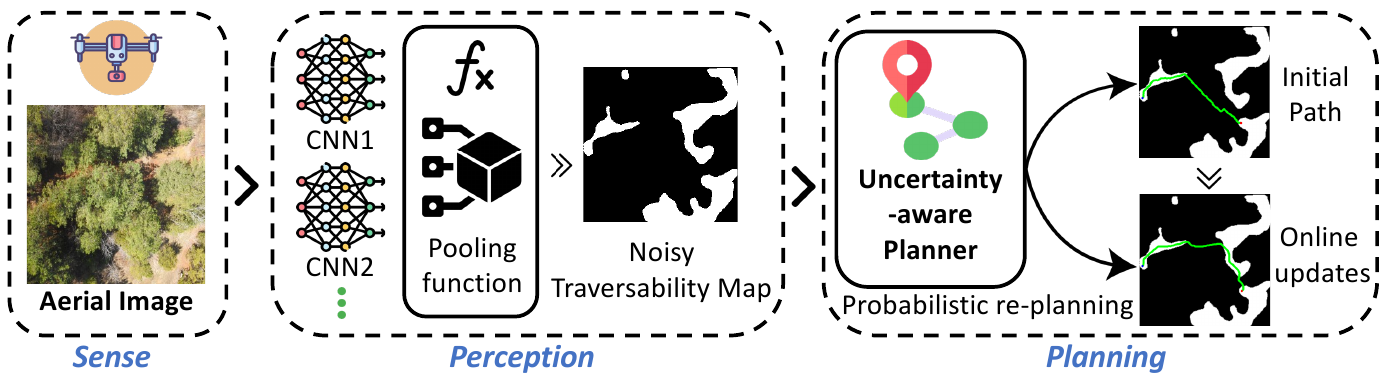}
  \caption{Proposed pipeline for uncertainty-aware perception and planning from aerial observations}
  \label{Fig:pipeline_diag}
  \vspace{-6mm}
\end{figure*}

\section{LITERATURE REVIEW}

Classical path planning algorithms in robotics mostly rely on static maps, which assume that information about which areas are traversable and which areas are not traversable are available in advance in the form of pre-built maps and do not change over time. Classical path planning may be further classified into sampling-based algorithms~\cite{sadat2020perceive} and search-based algorithms~\cite{ajanovic2018search}. Search-based algorithms include the popular Dijkstra's algorithm~\cite{dijkstra1959note}, A* algorithm~\cite{hart1968}, and state lattice algorithms~\cite{mcnaughton2011motion}.
Variants of search-based path planning include the weighted A* approach~\cite{pohl1970first}, which is faster and uses less memory, but is not optimal.
Alternatively, the Anytime Repairing A* (ARA*) algorithm~\cite{likhachev2003ara} provides a sub-optimal solution in a short period of time and continues to try to find an optimal solution within a specified time period.
More recent approaches have also used deep reinforcement learning~\cite{bhatia2022tuning} or Bayesian optimization~\cite{cano2018} to tune the hyperparameters of the planner.
On the other hand, sampling-based approaches such as RRT use random samples drawn from traversable areas of the search space, which allow planning to be carried out in non-convex and high-dimensional spaces~\cite{lavalle1998}\cite{van2021curvature}. Overall, classical path planning algorithms may work well in structured environments but fail to address the problem of unstructured off-road environments with complex terrain and uncertain traversability~\cite{hua2022}.

To make an informed decision regarding the desired path during autonomous navigation, it is essential to make use of real-time semantic information derived from sensors that are observing the surrounding environment~\cite{Cui_2021_ICCV}. Recent advances in computer vision~\cite{ranft2016} and the release of big datasets~\cite{liao2022kitti}\cite{cordts2016cityscapes} have facilitated research into path planning methods that can reason directly from sensor inputs~\cite{hu2023planning}\cite{Cui_2022_WACV}.
In the domain of ground images,~\cite{Can_2022_CVPR} proposes a neural network for predicting lane geometry and estimating a topology-preserving road network using a forward-looking camera. 
Similarly,~\cite{yuan2020segfix} uses neural networks to improve the boundary quality for road segmentation.
In the domain of aerial images, road semantic segmentation can also be carried out using convolutional neural networks (CNNs)~\cite{quan2021}\cite{yang2022sdunet}, or graph neural networks~\cite{Bahl_2022_CVPR}\cite{mei2021coanet} on UAV images or satellite images to provide traversability information to a ground vehicle.
Most of these existing works only focus on improving the traversability prediction task without implementing a path-planning solution, which can account for errors and uncertainty in the perception model output.

Another key research gap that we plan to address in this paper is path planning in off-road environments. Urban roads are characterized by features such as curbs, buildings, traffic signs, road markings, and guardrails that can simplify the perception and planning problem\cite{Volpi_2022_CVPR, li2022deep}, whereas rural roads lack clear boundaries and intersections are complex and heterogeneous~\cite{yadav2018rural, kearney2020maintaining}. The vast majority of autonomous driving systems have been trained using either urban or suburban datasets (e.g. KITTI~\cite{liao2022kitti} and Cityscapes~\cite{cordts2016cityscapes}) without consideration for rural environments. Some datasets such as Robot Unstructured Ground Driving (RUGD)~\cite{wigness2019rugd}, OFF-Road Freespace Detection (ORFD)~\cite{min2022orfd}, DeepScene~\cite{valada2016deep}, Center for Advanced Vehicular Systems Traversability (CaT)~\cite{sharma2022, carruth2022challenges}, do involve off-road environments but only evaluate perception tasks such as semantic segmentation and free-space detection and not planning tasks. In contrast, this paper will introduce a new aerial image dataset for off-road environments and use it as a benchmark for path planning.

\section{METHODOLOGY}

\subsection{Problem definition}
Given an aerial image, \textbf{I} of a region of interest, a start position, \textbf{s}, and goal position, \textbf{g}, in image coordinates: compute the \textit{best} path for a ground robot to travel from \textbf{s} to \textbf{g} using only information in \textbf{I}. The \textit{best} path is evaluated based on (i) quality, i.e. how short the total path length is for the computed path, and (ii) feasibility, i.e. how much of the computed path is actually traversable. \textbf{I} is assumed to be captured with an image plane parallel to the ground plane so that pixel-wise distances are roughly proportional to real-world physical distances.


\subsection{Traversable area segmentation from aerial images}
The segmentation step aims to take an aerial observation of a scene, pass it through a semantic information extractor in the form of a DNN, and finally predict the traversability of different regions in the scene at the individual pixel level~\cite{zhang2018road}. The output of the segmentation network for a given aerial image will be the traversal probability distribution matrix for that particular image.
For image segmentation, we utilize an ensemble of DNN methods to predict the traversal probabilities. Initially, the neural networks were pre-trained on the classification task using the ImageNet\cite{deng2009imagenet} dataset containing over 14 million images. Next, we fine-tuned the networks on the semantic segmentation task using specific aerial image datasets to enable the networks to perform traversability estimation from aerial images.

\subsubsection{Network architecture}
In this research, we developed an ensemble model utilizing output segmentation heads from U-Net~\cite{ronneberger2015u} and DeepLabV3+~\cite{chen2018encoder} built on a ResNet-50~\cite{he2016deep} encoder and pre-trained on ImageNet ~\cite{deng2009imagenet}. Upsampling layers from U-Net and atrous convolution layers from DeepLabV3 are both common strategies in image semantic segmentation to process multi-scale contextual information in image data. The segmentation heads are trained to predict binary traversability (either traversable or not-traversable for each pixel) using the Dice loss function~\cite{sudre2017generalised}. During inference, the output of the final softmax layer is used to extract a traversability map over the entire image. The middle dotted-line block in Figure \ref{Fig:pipeline_diag} shows the proposed network architecture for traversable areas from aerial images.

In our empirical studies, we found that having a high recall rate for traversable terrain is important for successfully generating paths from the start position to the goal position.
This is because if the ratio of regions predicted to be traversable compared to the regions predicted to be non-traversable is too low, the path planner may terminate prematurely before finding a traversable connection between the start position and the goal position.
Thus, the proposed network architecture uses a max-pooling layer to combine predictions from an ensemble of segmentation heads.
The output of the pooling layer has the highest probability of traversability among the input model predictions for each pixel.
In our experiments, we found that the pooling function is effective in achieving generally higher recall rates for traversable terrain (refer to Table \ref{table:segmentation_results} in the Results section). Theoretically, the outputs of more than two segmentation networks may be pooled together in the ensemble model; however, in our experiments, we found that pooling together two segmentation networks gave adequate performance.

\subsubsection{Aerial image datasets}

In this research, we make use of the Massachusetts Road Dataset (MRD)~\cite{MnihThesis} and the DeepGlobe dataset (DGD)~\cite{DeepGlobe18}, which are both datasets of satellite images with a mix of urban and off-road environments. 
MRD contained 1108 training, 49 testing, and 14 validation images, all of 1500 x1500 in resolution and with corresponding ground truth labels. DGD contained 6226 training, 1101 testing, and 1243 validation images, all of 1024 x1024 in resolution but with only the training set having ground truth labels.
We resized MRD and DGD images to a standard resolution of 1536 x1536 pixels to maintain consistency. Although these datasets are not directly applicable to the targeted domain of off-road environments, we used them for testing and comparison since these datasets are publicly released and have a large number of annotations readily available. For validation of the approach in the domain of off-road environments, we collected and annotated our own dataset of off-road aerial images obtained from the Center for Advanced Vehicular Systems (CAVS) proving grounds at Mississippi State University~\cite{trail_map} (hereafter referred to as the CAVS dataset).
The proving grounds is a 55-acre test facility featuring 12 rugged off-road trails filled with naturally occurring obstacles and terrain features such as rocks, tall grasses, wet lowlands, and wooded or obscured trails.
For our CAVS dataset, we manually labeled a total of 403 images and split them into training, test, and validation sets of 332, 38, and 33 images respectively.

In addition, we applied data augmentation to generate sufficiently diverse samples for training.
We pre-processed the images with random crop, horizontal flip, vertical flip, and random rotation at 0.75 probability for all the datasets.

\subsubsection{Hyperparameters}
The networks were trained for a total of 15 epochs with a batch size of 16. The Adam~\cite{kingma2014adam} optimizer was used due to its faster convergence and fewer hyperparameter requirements. Softmax was used as the activation function for the segmentation prediction layer. These hyperparameter settings follow widely used standard training procedures and have been previously applied on the MRD~\cite{unet_massachuests} and DeepGlobe datasets~\cite{deeplab_deepglobe}. Note that separate models were trained for each dataset.


\subsection{Uncertainty-aware path planning}
In this subsection, we introduce an Uncertainty-aware A* (URA*) approach to generate suitable paths with respect to uncertainty in unknown environments. In traditional A*-based approaches~\cite{hart1968}\cite{likhachev2003ara}\cite{ren2022}, the environment is first discretized into states, and searches over the state-space are carried out based on the edge costs as the optimality criteria. 
However, this is assuming a perfect environment where the traversability and cost of every state are known in advance. In this research, we take anytime-replanning techniques from ARA* \cite{likhachev2003ara} and extend it to uncertain environments by incorporating predictions from a semantic segmentation network to generate robust paths that take into account the traversal probability of each state. We utilize this URA* algorithm, in conjunction with D*-lite~\cite{Koenig2002Dlite} to extend to the replanning problem, with an algorithm we call Uncertainty-aware D*-lite (URD*) (described in the next subsection).

The traversability matrix obtained from the segmentation network is divided into a grid where each grid cell stores the traversal probability of that region. In this study, we resample the traversability matrix to a grid of 600x600 cells to speed up the computation. The path-planning algorithm will generate a sequence of cells to traverse from the start cell to the goal cell. A denser grid can be used to generate finer paths, at the cost of incurring higher computational time.

\begin{algorithm}
\small
\caption{URA* f-value}\label{alg:urafvalue}
\KwInput{Traversability Model Predictions, $M$, State, $s$}
\KwOutputput{f-value}
    return $g(s) + \epsilon * ( dist(s, goal) - (\alpha * M(s)) ) $
\end{algorithm}
\vspace{-5mm}



\begin{algorithm}
\small
\caption{URA*}
\label{alg:ura}
\KwInput{Traversability Model Predictions, $M$, $s_{start}$, $s_{goal}$}
\KwOutputput{Path from $s_{start}$ to $s_{goal}$ }
    $g(s_{start}) = \infty$; $g(s_{goal}) = 0$ \\
    $OPEN = CLOSED = INCONS = \varnothing$ \\
    Insert $s_{start}$ into $OPEN$ with URA\_f\_value($s_{start}$)\\
    ImprovePath()\\
    \While{$\epsilon > 1$}{
        Decrease $\epsilon$\\
        Move states from $INCONS$ into $OPEN$\\
        Update all priorities in $OPEN$ according to URA\_f\_value(s)\\
        $CLOSED = \varnothing$\\
        ImprovePath()
    }
\end{algorithm}





Algorithm \ref{alg:urafvalue} shows the f-value calculation of URA*, which determines the priority of which state to expand next.  Similar to weighted-A* and ARA*, an $\epsilon$ parameter is used to weight the heuristic vs. the g-value. The heuristic value for a state consists of the distance from the current state to the goal state subtracted by the traversal probability times a constant multiplier $\alpha$. This places a higher preference on nodes that have a higher probability of being free space and are also closer to the goal.

Algorithm \ref{alg:ura} shows the main loop of URA*. Similar to ARA*, this involves running weighted A* multiple times with $\epsilon$ gradually lowering each time. The $ImprovePath()$ function is borrowed from ARA* and recomputes the shortest path within a given $\epsilon$ while reusing search efforts from the previous executions. In $ImprovePath()$, the cost of visiting a node is calculated as $1-M(s)$; the higher the predicted traversability probability, the lower the cost of a state.





\subsection{Uncertainty-aware path replanning}

In this section, we introduce URD*, a probabilistic replanning technique that combines information from noisy aerial-to-ground traversability estimates with accurate ground-level traversability measurements. This algorithm is applied so that the ground robot is able to rapidly scan and re-plan suitable paths during physical operation. In order to effectively update the environment of the surrounding agent during traversal, we simulate LiDAR scans by using Bresenham's algorithm \cite{bresenham1965} to simulate the field of view of the robot as it moves through the environment and updates its internal representation of the traversable areas.


\begin{algorithm}
\label{alg:urd}
\caption{URD*}\label{algo:urd_*}
    InitializeEnvironment()\\
    Initialize Tree with URA*\\
    $s_{current} = s_{last} = s_{start}$\\
    \While{$s_{current} \neq goal$}{ 
        ComputeShortestPath($s_{current}$, $s_{goal}$)\\
        \If {$g(s_{current}) = \infty$}{
            return $fail$
        }
        $s_{current} = argmin_{s^{'} \epsilon Succ(s_{current})} (c(s_{current}, s^{'}) + g(s^{'}))$ \\
        Move to $s_{current}$ \\
        UpdateEnvironment() \\
        \If {any edge costs changed}{
            Update vertices with D*-lite procedure and URD* heuristic\\
        }      
    }    
\end{algorithm}

\subsubsection{Tree Initialization}
Using Algorithm \ref{alg:ura}, the search tree initialization step is performed with URA*. 
Since URA* uses traversability prediction values as pseudo-costs, the initial search process is guaranteed to always find a path from the start state to the goal state.









\begin{algorithm}
\label{alg:urdheuristic}
\small
\caption{URD* heuristic}
\KwInput{Traversability Model Predictions, $M$, $s_{start}$, $s_{current}$}
\KwOutputput{Heuristic value of $s$}

    $d_{x} = || x_{start}-x_{current} ||$\\
    $d_{y} = || y_{start}-y_{current} || $\\
    
    
    $u(s) = dist(s_{start}, s) / dist(s_{start}, s_{goal})$ \\
    $H = u(s) * (\gamma * min(d_x, d_y) + || d_x – d_y ||) $ \\
    return $min(H, dist(s_{start}, s_{current}))$\\
\end{algorithm}

\subsubsection{Replanning} In Algorithm \ref{alg:urd}, we adopt similar procedures to D*-lite \cite{Koenig2002Dlite} to replan paths to the goal, starting from the initial URA* search tree. Each time the simulated robot moves to a new state, $s_{current}$, the traversability costs of the environment is updated by scanning a fixed radius around the robot and assigning the true traversability (i.e. the ground truth labels in the MRD, DeepGlobe, and CAVS datasets). This is implemented in the function $UpdateEnvironment()$. Then, if the edge costs have changed, the vertices are updated according to the D*-lite procedure in combination with our new URD* heuristic.

\subsubsection{Improved Heuristic}
In Algorithm \ref{alg:urdheuristic}, we determine the best node to expand during the replanning process by establishing a custom heuristic. We place higher importance on nodes that have a high traversal probability score as generated from the segmentation model and are closer to the goal. This heuristic is similar to the heuristic presented in \cite{9327404}, and we utilize a similar calculation method with a $u(s)$ value term in the heuristic $H$, where $\gamma$ is the constant multiplier. We place a weight on the first equation to bias the algorithm away from the customized heuristic toward a simpler Euclidean distance heuristic as the number of replans increases as this indicates overestimation. 

\subsubsection{Tree Resetting and Heuristic Scaling}
In order to prevent the algorithm from being trapped in deadlock situations, we reset the search tree and plan a new path to the goal whenever $s_{current}$ did not update after a few iterations. We also scale the $\gamma$ term after each replan, to resort to the Euclidean distance term in the event that the segmentation model is highly inaccurate.

\section{RESULTS}

\subsection{Performance analysis of traversability segmentation}

\begin{figure*}[h]
  \centering
  \includegraphics[scale=0.5]{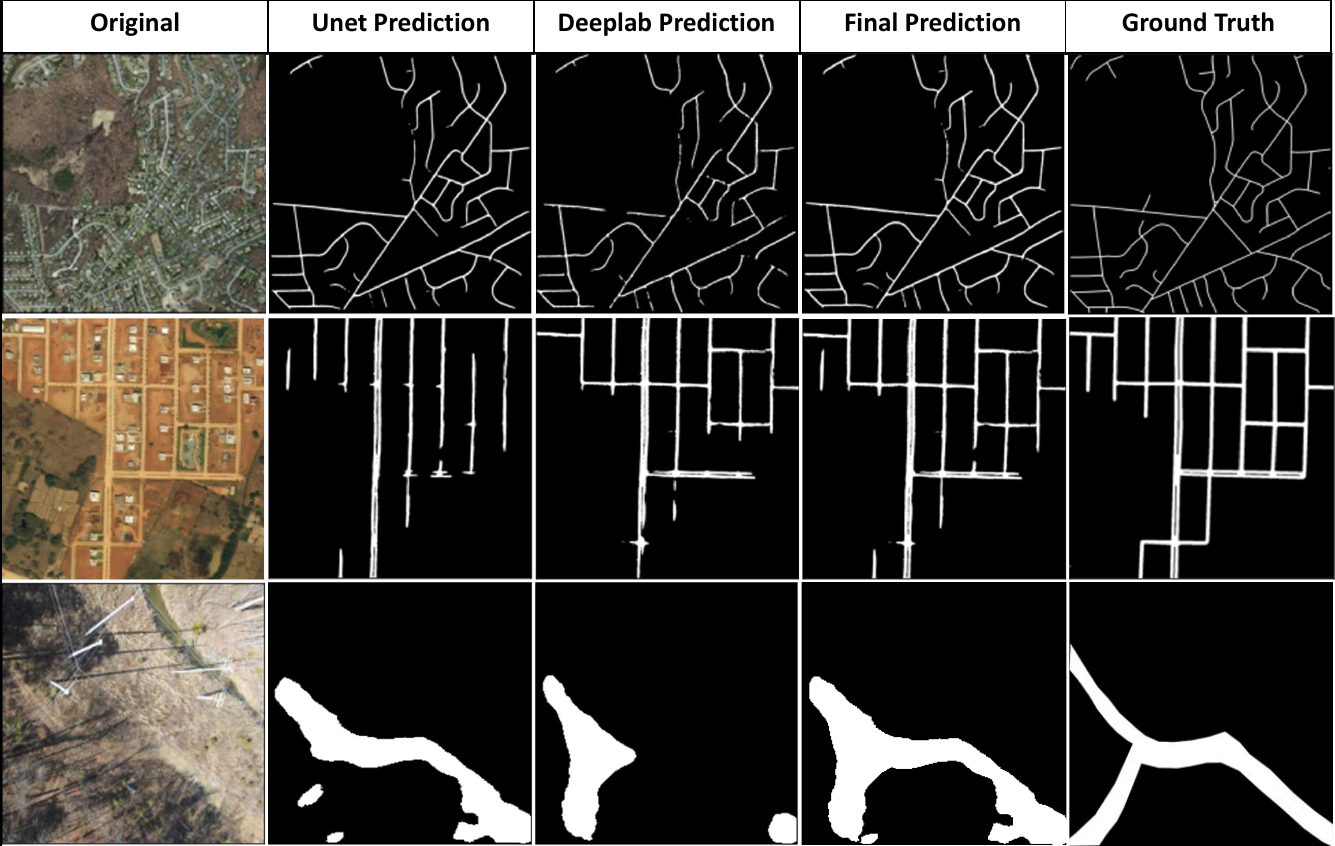}
  \caption{Traversability segmentation results for aerial images from different datasets. From top to bottom, the rows represent images from the Massachusetts Road Dataset, DeepGlobe dataset, and CAVS dataset. From left to right, the columns represent the (i) original image, (ii) predicted segmentation mask PSM from  U-Net (iii) PSM from  DeepLabV3+ (iv) PSM from our ensemble model, and (v) ground truth segmentation mask}
  \label{Fig:prediction_diag}
\end{figure*}

\begin{figure*}[h]
  \centering
  \includegraphics[scale=0.42]{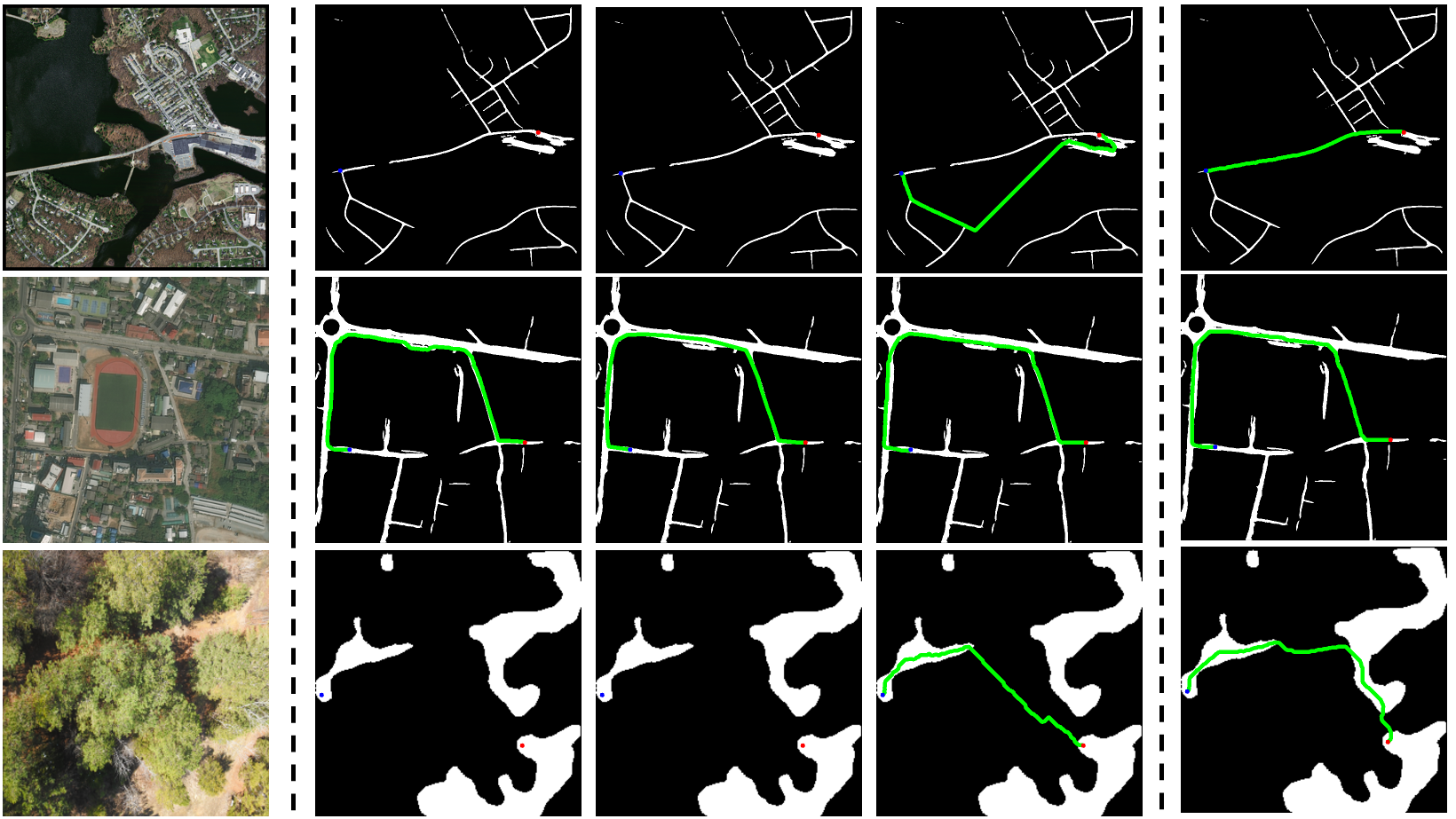}
  \caption{Path planning results for aerial images from different datasets. From top to bottom, the rows represent images from the Massachusetts Road Dataset, DeepGlobe dataset, and CAVS dataset. From left to right, the columns represent the (i) input aerial image, (ii) A* path (iii) RRT* path (iv) proposed URA* path and (v) proposed URD* replanned path. Red/blue dots indicate start/goal points whereas green lines indicate planned path. A path is not plotted if the algorithm fails to find a path between the start and goal points.}
  \label{Fig:path_diag}
\end{figure*}

Table \ref{table:segmentation_results} shows the segmentation performance analysis of UNet, DeepLabV3+, and our ensemble model on the MRD, DeepGlobe, and CAVS datasets. The traversability predictions were compared pixel-by-pixel to the ground truth annotations. The evaluation metrics used are Dice Loss, standard deviation (SD) of Dice Loss, Intersection-over-Union (IoU), and recall rate, averaged over all images for each dataset. Bold numbers indicate the best-performing model for each metric.

Results show that the proposed ensemble model for traversability segmentation achieved the lowest standard deviation in Dice Loss and the highest IoU for two out of the three datasets. More importantly, the proposed model achieved the highest recall rate for all datasets, demonstrating the benefit of an ensemble approach for maximizing the rate of finding traversable regions from aerial images. Still, the recall rates remain low at 50-60\% across the three datasets. In the next section, we will present our results of uncertainty-aware reasoning and path planning to overcome these noisy segmentation results.

\begin{table}[h]
  \caption{Segmentation performance comparison}
  \label{table:segmentation_results}
  \centering
  \resizebox{\columnwidth}{!}{
  \begin{tabular}{cccccc}
    \toprule
    Dataset & Method & Dice Loss & Dice\_SD & IoU(\%) & Recall\\
    \midrule
MRD & UNet & 0.1820 & 0.068 & 65.15 & 0.47\\
& DeepLabV3+ & \textbf{0.0637} & 0.066 & 62.58 & 0.37\\
& Ensemble(ours) & 0.1558 & \textbf{0.065} & \textbf{65.21} & \textbf{0.50}\\
\midrule
DeepGlobe & UNet & 0.2607 & 0.080 & 74.09 & 0.60\\
& DeepLabV3+ & \textbf{0.1773} & 0.103 & 65.30 & 0.36\\
& Ensemble(ours) & 0.2647 & \textbf{0.079} & \textbf{74.51} & \textbf{0.62}\\
\midrule
CAVS & UNet & 0.2069 & 0.102 & 62.68 & 0.51\\
& DeepLabV3+ & \textbf{0.1473} & 0.102 & \textbf{56.78} & 0.33\\
& Ensemble(ours) & 0.2049 & \textbf{0.093} & 61.93 & \textbf{0.55}\\
    \bottomrule
  \end{tabular}
  }
\vspace{-5mm}
\end{table}

\subsection{Performance analysis of path planning}
To evaluate the performance of URA* for calculation of the initial path, we opted to compare the algorithm with RRT* and A*, which are popular algorithms for path planning. For A* and RRT*, a confidence threshold of 50\% was used as the cutoff threshold for converting the segmentation network predictions to a binary traversability map. We also considered an alternate version of A*, which we term as A**, where we lower the confidence threshold from 50\% to 30\% to give it a better chance of obtaining an initial solution. RRT* uses a step size of 5, search radius of 50, and 10000 iterations. In the path planning experiments, we manually fix the start and end points for each aerial image.

The results of path planning for generating the initial path are shown in Table \ref{table:initial_planning}. These results are obtained from 49 images in the MRD test set, 29 images in the DeepGlobe validation set (since ground truth labels for the DeepGlobe test set has not been released), and 38 images in the CAVS test set. We use the normalized path length, average path accuracy, and success rate as evaluation metrics. The path length reflects the \textit{quality} of the planned path whereas the path accuracy reflects the \textit{feasibility} of the planned path. The normalized path length is calculated by dividing the computed path length in pixels with the straight line distance from start to goal in pixels. The path accuracy is calculated by comparing the pixels of the computed path with the ground truth traversability labels to determine the percentage of the computed path that lie in traversable regions. That is, the higher the path accuracy, the more likely the initial path is to be feasible for a robot. Finally, the success rate is calculated as the percentage of aerial images for which the path planning algorithm is able to generate an initial path without returning failure. Note that in cases where an algorithm is not successful in producing a complete path from the start to goal (given an input image), we use the maximum cost computed among all algorithms as a nominal value to penalize these failure cases.

Results in Table \ref{table:initial_planning} show that the proposed URA* algorithm significantly outperforms baseline algorithms on normalized path length, path accuracy and success rate but expands significantly more nodes than A* to find a feasible solution. By integrating the traversability probabilities into the planning process, URA* is able to generate higher quality and more feasible paths. In addition, URA* is always successful in returning a solution. In contrast to A* or RRT*, which treats the input map as having binary traversability and may terminate prematurely if there are insufficient areas predicted to be traversable, URA* is designed to always be able to obtain a path from the start to goal by treating traversability as a continuous value.


The results of path planning for generating replanned paths for online operations are shown in Table \ref{table:replanning}. Rapidly-replanning A* (RRA*) \cite{ganganath2016} and D*-lite \cite{Koenig2002Dlite} are used as baseline algorithms. Results show that the proposed URD* algorithm performs the best in terms of shortest path length and fewest nodes expanded in two out of three datasets considered. For the CAVS dataset, URD* performs slightly worse compared to RRA* because the dataset contains scenes with fewer twists and intersections and thus, the advantage of uncertainty-aware replanning was not as significant compared to the MRD and DeepGlobe datasets.

Figure \ref{Fig:path_diag} shows a visual comparison of the paths generated by the proposed algorithm overlaid on the predicted traversability maps. Results show that A* and RRT* mostly fail or take suboptimal paths due to the noisy traversability maps whereas URA* is able to generate reasonable paths and URD* can improve those paths after replanning.

\begin{table}[h]
  \caption{Performance comparison of the initial generated path}
  \label{table:initial_planning}
  \centering
  \resizebox{\columnwidth}{!}{
  \begin{tabular}{cccccc}
    \toprule
    \makecell{Dataset}&Method&\makecell{Norm. Path\\Length}&Path Acc.(\%) &\makecell{Success\\ Rate(\%)} &\makecell{Avg.\\Nodes\\Expanded} \\
    \midrule
MRD & A* & 1.93 & 11.89 & 44.90 & 1190\\
& A** & 1.93 & 12.67 & 44.90 & \textbf{1123}\\ 
& RRT* & 1.96 & 26.14 & 40.82 & 10000\\
&\makecell{URA*} & \textbf{1.25} & \textbf{44.70} & \textbf{100.0} & 4192\\
\midrule
DeepGlobe & A* & 2.31 & 34.75 & 46.43 & \textbf{6341}\\
& A** & 2.31 & 35.94 & 46.43 & 6368\\
& RRT* & 2.30 &  40.36 & 46.43 & 10000\\
&\makecell{URA*} & \textbf{1.52} & \textbf{83.37} & \textbf{100.0} & 12842\\
\midrule
CAVS & A* & 2.10 & 14.95 & 21.05 & \textbf{8181}\\
& A** & 2.07 & 16.30 & 23.68 & 8648 \\
& RRT* & 2.09 & 19.42 & 21.05 & 10000 \\
& \makecell{URA*} & \textbf{1.17} & \textbf{85.01} & \textbf{100.0} & 30762 \\ 
    \bottomrule
\end{tabular}}
\vspace{-3mm}
\end{table}

\begin{table}[h]
  \caption{Performance comparison of replanned path}
  \label{table:replanning}
  \centering
  \begin{tabular}{cccc}
    \toprule
    Dataset & Method &\makecell{Norm. Path\\Length}& \makecell{Avg. Nodes\\Expanded}\\
    \midrule
    MRD & RRA* & 2.32 & 346\\
    & D*-lite & 2.01 & 472\\
    & URD* & \textbf{1.36} & \textbf{300}\\
    \midrule
    DeepGlobe & RRA* & 2.33 & 505\\
    & D*-lite & 2.16 & 664\\
    & URD* & \textbf{1.50} & \textbf{490}\\
    \midrule
    CAVS & RRA* & \textbf{1.21} & \textbf{161}\\
    & D*-lite & 1.40 & 359\\
    & URD* & 1.24 & 262\\
    \bottomrule
  \end{tabular}
  \vspace{-5mm}
\end{table}

\section{CONCLUSIONS}
In conclusion, this research demonstrated an uncertainty-aware path planning algorithm to compute the best path through a region with unknown traversability where only aerial images are available.
In future work, we will investigate the possibility of using real-time traversability observations to update the segmentation network model to generate more accurate traversability estimations for replanning purposes. We will also conduct experiments with off-road vehicles to benchmark the effectiveness of this form of aerial-to-ground traversability estimation and planning in real-world conditions.


\section*{ACKNOWLEDGMENT}

The work reported herein was supported by by the National Science Foundation (NSF) (Award \#IIS-2153101). Any opinions, findings, and conclusions or recommendations expressed in this material are those of the authors and do not necessarily reflect the views of the NSF.


{\small
\bibliographystyle{IEEEtran}

\bibliography{main}

\begin{thebibliography}{10}
\providecommand{\url}[1]{#1}
\csname url@samestyle\endcsname
\providecommand{\newblock}{\relax}
\providecommand{\bibinfo}[2]{#2}
\providecommand{\BIBentrySTDinterwordspacing}{\spaceskip=0pt\relax}
\providecommand{\BIBentryALTinterwordstretchfactor}{4}
\providecommand{\BIBentryALTinterwordspacing}{\spaceskip=\fontdimen2\font plus
\BIBentryALTinterwordstretchfactor\fontdimen3\font minus
  \fontdimen4\font\relax}
\providecommand{\BIBforeignlanguage}[2]{{%
\expandafter\ifx\csname l@#1\endcsname\relax
\typeout{** WARNING: IEEEtran.bst: No hyphenation pattern has been}%
\typeout{** loaded for the language `#1'. Using the pattern for}%
\typeout{** the default language instead.}%
\else
\language=\csname l@#1\endcsname
\fi
#2}}
\providecommand{\BIBdecl}{\relax}
\BIBdecl

\bibitem{borges2022}
P.~V.~K. Borges, T.~Peynot, S.~Liang, B.~Arain, M.~Wildie, M.~G. Minareci,
  S.~Lichman, G.~Samvedi, I.~Sa, N.~Hudson, M.~Milford, P.~Moghadam, and
  P.~Corke, ``A survey on terrain traversability analysis for autonomous ground
  vehicles: Methods, sensors, and challenges,'' \emph{Field Robotics}, vol.~2,
  pp. 1567--1627, 2022.

\bibitem{sharma2022}
S.~Sharma, L.~Dabbiru, T.~Hannis, G.~Mason, D.~W. Carruth, M.~Doude, C.~Goodin,
  C.~Hudson, S.~Ozier, J.~E. Ball, and B.~Tang, ``Cat: Cavs traversability
  dataset for off-road autonomous driving,'' \emph{IEEE Access}, vol.~10, pp.
  24\,759--24\,768, 2022.

\bibitem{levi2015}
D.~Levi, N.~Garnett, and E.~Fetaya, ``Stixelnet: A deep convolutional network
  for obstacle detection and road segmentation,'' in \emph{Proceedings of the
  British Machine Vision Conference (BMVC)}.\hskip 1em plus 0.5em minus
  0.4em\relax BMVA Press, September 2015, pp. 109.1--109.12.

\bibitem{oliveira2016}
G.~L. Oliveira, W.~Burgard, and T.~Brox, ``Efficient deep models for monocular
  road segmentation,'' in \emph{IEEE/RSJ International Conference on
  Intelligent Robots and Systems (IROS)}, 2016, pp. 4885--4891.

\bibitem{chavez2018}
R.~O. Chavez-Garcia, J.~Guzzi, L.~M. Gambardella, and A.~Giusti, ``Learning
  ground traversability from simulations,'' \emph{IEEE Robotics and Automation
  Letters}, vol.~3, no.~3, pp. 1695--1702, 2018.

\bibitem{kim2019}
P.~Kim, J.~Park, Y.~K. Cho, and J.~Kang, ``Uav-assisted autonomous mobile robot
  navigation for as-is 3d data collection and registration in cluttered
  environments,'' \emph{Automation in Construction}, vol. 106, p. 102918, 2019.

\bibitem{hudjakov2009}
R.~Hudjakov and M.~Tamre, ``Aerial imagery terrain classification for
  long-range autonomous navigation,'' in \emph{2009 International Symposium on
  Optomechatronic Technologies}, 2009, pp. 88--91.

\bibitem{bandara2021}
W.~G.~C. Bandara, J.~M.~J. Valanarasu, and V.~M. Patel, ``{SPIN} road mapper:
  Extracting roads from aerial images via spatial and interaction space graph
  reasoning for autonomous driving,'' \emph{CoRR}, vol. abs/2109.07701, 2021.

\bibitem{quan2021}
B.~Quan, B.~Liu, D.~Fu, H.~Chen, and X.~Liu, ``Improved deeplabv3 for better
  road segmentation in remote sensing images,'' in \emph{2021 International
  Conference on Computer Engineering and Artificial Intelligence (ICCEAI)},
  2021, pp. 331--334.

\bibitem{hosseinpoor2021}
S.~Hosseinpoor, J.~Torresen, M.~Mantelli, D.~Pitto, M.~Kolberg, R.~Maffei, and
  E.~Prestes, ``Traversability analysis by semantic terrain segmentation for
  mobile robots,'' in \emph{IEEE 17th International Conference on Automation
  Science and Engineering (CASE)}, 2021.

\bibitem{lavalle1998}
S.~M. LaValle, ``Rapidly-exploring random trees : a new tool for path
  planning,'' \emph{Technical Report. Computer Science Department, Iowa State
  University}, 1998.

\bibitem{hart1968}
P.~E. Hart, N.~J. Nilsson, and B.~Raphael, ``A formal basis for the heuristic
  determination of minimum cost paths,'' \emph{IEEE Transactions on Systems
  Science and Cybernetics}, vol.~4, no.~2, pp. 100--107, 1968.

\bibitem{ono2015}
M.~Ono, T.~J. Fuchs, A.~Steffy, M.~Maimone, and J.~Yen, ``Risk-aware planetary
  rover operation: Autonomous terrain classification and path planning,'' in
  \emph{2015 IEEE Aerospace Conference}, 2015, pp. 1--10.

\bibitem{candela2022}
A.~Candela and D.~Wettergreen, ``An approach to science and risk-aware
  planetary rover exploration,'' \emph{IEEE Robotics and Automation Letters},
  vol.~7, no.~4, pp. 9691--9698, 2022.

\bibitem{sadat2020perceive}
A.~Sadat, S.~Casas, M.~Ren, X.~Wu, P.~Dhawan, and R.~Urtasun, ``Perceive,
  predict, and plan: Safe motion planning through interpretable semantic
  representations,'' in \emph{European Conference on Computer Vision}.\hskip
  1em plus 0.5em minus 0.4em\relax Springer, 2020, pp. 414--430.

\bibitem{ajanovic2018search}
Z.~Ajanovic, B.~Lacevic, B.~Shyrokau, M.~Stolz, and M.~Horn, ``Search-based
  optimal motion planning for automated driving,'' in \emph{2018 IEEE/RSJ
  International Conference on Intelligent Robots and Systems (IROS)}.\hskip 1em
  plus 0.5em minus 0.4em\relax IEEE, 2018, pp. 4523--4530.

\bibitem{dijkstra1959note}
E.~Dijkstra, ``A note on two problems in connection with graphs,''
  \emph{Numerische Mathematik}, vol.~1, no.~1, pp. 269--271, 1959.

\bibitem{mcnaughton2011motion}
M.~McNaughton, C.~Urmson, J.~M. Dolan, and J.-W. Lee, ``Motion planning for
  autonomous driving with a conformal spatiotemporal lattice,'' in \emph{2011
  IEEE International Conference on Robotics and Automation}.\hskip 1em plus
  0.5em minus 0.4em\relax IEEE, 2011, pp. 4889--4895.

\bibitem{pohl1970first}
I.~Pohl, ``First results on the effect of error in heuristic search,''
  \emph{Machine Intelligence}, vol.~5, pp. 219--236, 1970.

\bibitem{likhachev2003ara}
M.~Likhachev, G.~J. Gordon, and S.~Thrun, ``Ara*: Anytime a* with provable
  bounds on sub-optimality,'' \emph{Advances in neural information processing
  systems}, vol.~16, 2003.

\bibitem{bhatia2022tuning}
A.~Bhatia, J.~Svegliato, S.~B. Nashed, and S.~Zilberstein, ``Tuning the
  hyperparameters of anytime planning: A metareasoning approach with deep
  reinforcement learning,'' in \emph{Proceedings of the International
  Conference on Automated Planning and Scheduling}, vol.~32, 2022, pp.
  556--564.

\bibitem{cano2018}
J.~Cano, Y.~Yang, B.~Bodin, V.~Nagarajan, and M.~O'Boyle, ``Automatic parameter
  tuning of motion planning algorithms,'' in \emph{2018 IEEE/RSJ International
  Conference on Intelligent Robots and Systems (IROS)}, 2018, pp. 8103--8109.

\bibitem{van2021curvature}
B.~van~den Berg, B.~Brito, M.~Alirezaei, and J.~Alonso-Mora, ``Curvature aware
  motion planning with closed-loop rapidly-exploring random trees,'' in
  \emph{2021 IEEE Intelligent Vehicles Symposium (IV)}.\hskip 1em plus 0.5em
  minus 0.4em\relax IEEE, 2021, pp. 1024--1030.

\bibitem{hua2022}
\BIBentryALTinterwordspacing
C.~Hua, R.~Niu, B.~Yu, X.~Zheng, R.~Bai, and S.~Zhang, ``A global path planning
  method for unmanned ground vehicles in off-road environments based on
  mobility prediction,'' \emph{Machines}, vol.~10, no.~5, 2022. [Online].
  Available: \url{https://www.mdpi.com/2075-1702/10/5/375}
\BIBentrySTDinterwordspacing

\bibitem{Cui_2021_ICCV}
A.~Cui, S.~Casas, A.~Sadat, R.~Liao, and R.~Urtasun, ``Lookout: Diverse
  multi-future prediction and planning for self-driving,'' in \emph{Proceedings
  of the IEEE/CVF International Conference on Computer Vision (ICCV)}, October
  2021, pp. 16\,107--16\,116.

\bibitem{ranft2016}
B.~Ranft and C.~Stiller, ``The role of machine vision for intelligent
  vehicles,'' \emph{IEEE Transactions on Intelligent Vehicles}, vol.~1, no.~1,
  pp. 8--19, 2016.

\bibitem{liao2022kitti}
Y.~Liao, J.~Xie, and A.~Geiger, ``Kitti-360: A novel dataset and benchmarks for
  urban scene understanding in 2d and 3d,'' \emph{IEEE Transactions on Pattern
  Analysis and Machine Intelligence}, 2022.

\bibitem{cordts2016cityscapes}
M.~Cordts, M.~Omran, S.~Ramos, T.~Rehfeld, M.~Enzweiler, R.~Benenson,
  U.~Franke, S.~Roth, and B.~Schiele, ``The cityscapes dataset for semantic
  urban scene understanding,'' in \emph{Proceedings of the IEEE conference on
  computer vision and pattern recognition}, 2016, pp. 3213--3223.

\bibitem{hu2023planning}
Y.~Hu, J.~Yang, L.~Chen, K.~Li, C.~Sima, X.~Zhu, S.~Chai, S.~Du, T.~Lin,
  W.~Wang \emph{et~al.}, ``Planning-oriented autonomous driving,'' in
  \emph{Proceedings of the IEEE/CVF Conference on Computer Vision and Pattern
  Recognition}, 2023, pp. 17\,853--17\,862.

\bibitem{Cui_2022_WACV}
Y.~Cui, Z.~Cao, Y.~Xie, X.~Jiang, F.~Tao, Y.~V. Chen, L.~Li, and D.~Liu,
  ``Dg-labeler and dgl-mots dataset: Boost the autonomous driving perception,''
  in \emph{Proceedings of the IEEE/CVF Winter Conference on Applications of
  Computer Vision (WACV)}, January 2022, pp. 58--67.

\bibitem{Can_2022_CVPR}
Y.~B. Can, A.~Liniger, D.~P. Paudel, and L.~Van~Gool, ``Topology preserving
  local road network estimation from single onboard camera image,'' in
  \emph{Proceedings of the IEEE/CVF Conference on Computer Vision and Pattern
  Recognition (CVPR)}, June 2022, pp. 17\,263--17\,272.

\bibitem{yuan2020segfix}
Y.~Yuan, J.~Xie, X.~Chen, and J.~Wang, ``Segfix: Model-agnostic boundary
  refinement for segmentation,'' in \emph{European Conference on Computer
  Vision}.\hskip 1em plus 0.5em minus 0.4em\relax Springer, 2020, pp. 489--506.

\bibitem{yang2022sdunet}
M.~Yang, Y.~Yuan, and G.~Liu, ``Sdunet: Road extraction via spatial enhanced
  and densely connected unet,'' \emph{Pattern Recognition}, vol. 126, p.
  108549, 2022.

\bibitem{Bahl_2022_CVPR}
G.~Bahl, M.~Bahri, and F.~Lafarge, ``Single-shot end-to-end road graph
  extraction,'' in \emph{Proceedings of the IEEE/CVF Conference on Computer
  Vision and Pattern Recognition (CVPR) Workshops}, June 2022, pp. 1403--1412.

\bibitem{mei2021coanet}
J.~Mei, R.-J. Li, W.~Gao, and M.-M. Cheng, ``Coanet: Connectivity attention
  network for road extraction from satellite imagery,'' \emph{IEEE Transactions
  on Image Processing}, vol.~30, pp. 8540--8552, 2021.

\bibitem{Volpi_2022_CVPR}
R.~Volpi, P.~De~Jorge, D.~Larlus, and G.~Csurka, ``On the road to online
  adaptation for semantic image segmentation,'' in \emph{Proceedings of the
  IEEE/CVF Conference on Computer Vision and Pattern Recognition (CVPR)}, June
  2022, pp. 19\,184--19\,195.

\bibitem{li2022deep}
L.~Li, T.~Zhou, W.~Wang, J.~Li, and Y.~Yang, ``Deep hierarchical semantic
  segmentation,'' in \emph{Proceedings of the IEEE/CVF Conference on Computer
  Vision and Pattern Recognition}, 2022, pp. 1246--1257.

\bibitem{yadav2018rural}
M.~Yadav and A.~K. Singh, ``Rural road surface extraction using mobile lidar
  point cloud data,'' \emph{Journal of the Indian Society of Remote Sensing},
  vol.~46, no.~4, pp. 531--538, 2018.

\bibitem{kearney2020maintaining}
S.~P. Kearney, N.~C. Coops, S.~Sethi, and G.~B. Stenhouse, ``Maintaining
  accurate, current, rural road network data: An extraction and updating
  routine using rapideye, participatory gis and deep learning,''
  \emph{International Journal of Applied Earth Observation and Geoinformation},
  vol.~87, p. 102031, 2020.

\bibitem{wigness2019rugd}
M.~Wigness, S.~Eum, J.~G. Rogers, D.~Han, and H.~Kwon, ``A rugd dataset for
  autonomous navigation and visual perception in unstructured outdoor
  environments,'' in \emph{2019 IEEE/RSJ International Conference on
  Intelligent Robots and Systems (IROS)}.\hskip 1em plus 0.5em minus
  0.4em\relax IEEE, 2019, pp. 5000--5007.

\bibitem{min2022orfd}
C.~Min, W.~Jiang, D.~Zhao, J.~Xu, L.~Xiao, Y.~Nie, and B.~Dai, ``Orfd: A
  dataset and benchmark for off-road freespace detection,'' in \emph{2022
  International Conference on Robotics and Automation (ICRA)}.\hskip 1em plus
  0.5em minus 0.4em\relax IEEE, 2022, pp. 2532--2538.

\bibitem{valada2016deep}
A.~Valada, G.~L. Oliveira, T.~Brox, and W.~Burgard, ``Deep multispectral
  semantic scene understanding of forested environments using multimodal
  fusion,'' in \emph{International symposium on experimental robotics}.\hskip
  1em plus 0.5em minus 0.4em\relax Springer, 2016, pp. 465--477.

\bibitem{carruth2022challenges}
D.~W. Carruth, C.~T. Walden, C.~Goodin, and S.~C. Fuller, ``Challenges in low
  infrastructure and off-road automated driving,'' in \emph{2022 Fifth
  International Conference on Connected and Autonomous Driving
  (MetroCAD)}.\hskip 1em plus 0.5em minus 0.4em\relax IEEE, 2022, pp. 13--20.

\bibitem{zhang2018road}
Z.~Zhang, Q.~Liu, and Y.~Wang, ``Road extraction by deep residual u-net,''
  \emph{IEEE Geoscience and Remote Sensing Letters}, vol.~15, no.~5, pp.
  749--753, 2018.

\bibitem{deng2009imagenet}
J.~Deng, W.~Dong, R.~Socher, L.-J. Li, K.~Li, and L.~Fei-Fei, ``Imagenet: A
  large-scale hierarchical image database,'' in \emph{2009 IEEE conference on
  computer vision and pattern recognition}.\hskip 1em plus 0.5em minus
  0.4em\relax Ieee, 2009, pp. 248--255.

\bibitem{ronneberger2015u}
O.~Ronneberger, P.~Fischer, and T.~Brox, ``U-net: Convolutional networks for
  biomedical image segmentation,'' in \emph{International Conference on Medical
  image computing and computer-assisted intervention}.\hskip 1em plus 0.5em
  minus 0.4em\relax Springer, 2015, pp. 234--241.

\bibitem{chen2018encoder}
L.-C. Chen, Y.~Zhu, G.~Papandreou, F.~Schroff, and H.~Adam, ``Encoder-decoder
  with atrous separable convolution for semantic image segmentation,'' in
  \emph{Proceedings of the European conference on computer vision (ECCV)},
  2018, pp. 801--818.

\bibitem{he2016deep}
K.~He, X.~Zhang, S.~Ren, and J.~Sun, ``Deep residual learning for image
  recognition,'' in \emph{Proceedings of the IEEE conference on computer vision
  and pattern recognition}, 2016, pp. 770--778.

\bibitem{sudre2017generalised}
C.~H. Sudre, W.~Li, T.~Vercauteren, S.~Ourselin, and M.~Jorge~Cardoso,
  ``Generalised dice overlap as a deep learning loss function for highly
  unbalanced segmentations,'' in \emph{Deep Learning in Medical Image Analysis
  and Multimodal Learning for Clinical Decision Support: Third International
  Workshop, DLMIA 2017, and 7th International Workshop, ML-CDS 2017, Held in
  Conjunction with MICCAI 2017, Qu{\'e}bec City, QC, Canada, September 14,
  Proceedings 3}.\hskip 1em plus 0.5em minus 0.4em\relax Springer, 2017, pp.
  240--248.

\bibitem{MnihThesis}
V.~Mnih, ``Machine learning for aerial image labeling,'' Ph.D. dissertation,
  University of Toronto, 2013.

\bibitem{DeepGlobe18}
I.~Demir, K.~Koperski, D.~Lindenbaum, G.~Pang, J.~Huang, S.~Basu, F.~Hughes,
  D.~Tuia, and R.~Raskar, ``Deepglobe 2018: A challenge to parse the earth
  through satellite images,'' in \emph{The IEEE Conference on Computer Vision
  and Pattern Recognition (CVPR) Workshops}, 2018.

\bibitem{trail_map}
\BIBentryALTinterwordspacing
``Cavs proving ground map,'' 2022. [Online]. Available:
  \url{https://www.cavs.msstate.edu/capabilities/proving_ground.php}
\BIBentrySTDinterwordspacing

\bibitem{kingma2014adam}
D.~P. Kingma and J.~Ba, ``Adam: A method for stochastic optimization,''
  \emph{arXiv preprint arXiv:1412.6980}, 2014.

\bibitem{unet_massachuests}
\BIBentryALTinterwordspacing
B.~ASHWATH, ``Unet (resnet50 frontend) road segmentation pytorch,'' 2020.
  [Online]. Available:
  \url{https://kaggle.com/code/balraj98/unet-resnet50-frontend-road-segmentation-pytorch}
\BIBentrySTDinterwordspacing

\bibitem{deeplab_deepglobe}
\BIBentryALTinterwordspacing
B.~Ashwath, ``Road extraction from satellite images [deeplabv3+],'' 2020.
  [Online]. Available:
  \url{https://www.kaggle.com/code/balraj98/road-extraction-from-satellite-images-deeplabv3}
\BIBentrySTDinterwordspacing

\bibitem{ren2022}
Z.~Ren, S.~Rathinam, M.~Likhachev, and H.~Choset, ``Multi-objective path-based
  d* lite,'' \emph{IEEE Robotics and Automation Letters}, vol.~7, no.~2, pp.
  3318--3325, 2022.

\bibitem{Koenig2002Dlite}
\BIBentryALTinterwordspacing
S.~Koenig and M.~Likhachev, ``D*lite,'' in \emph{AAAI/IAAI}, 2002. [Online].
  Available: \url{https://api.semanticscholar.org/CorpusID:208940224}
\BIBentrySTDinterwordspacing

\bibitem{bresenham1965}
J.~E. Bresenham, ``Algorithm for computer control of a digital plotter,''
  \emph{IBM Systems Journal}, vol.~4, no.~1, pp. 25--30, 1965.

\bibitem{9327404}
J.~Yu, G.~Liu, Z.~Zhao, X.~Wang, J.~Xu, and Y.~Bai, ``Improved d*lite algorithm
  path planning in complex environment,'' in \emph{2020 Chinese Automation
  Congress (CAC)}, 2020, pp. 2226--2230.

\bibitem{ganganath2016}
N.~Ganganath, C.-T. Cheng, and C.~K. Tse, ``Rapidly replanning a*,'' in
  \emph{2016 International Conference on Cyber-Enabled Distributed Computing
  and Knowledge Discovery (CyberC)}, 2016, pp. 386--389.

\end{thebibliography}
}

\end{document}